\documentclass[journal]{IEEEtran}
\usepackage{algorithm}
\usepackage{algorithmic}
\usepackage{graphicx}
\usepackage{caption,subcaption}
\usepackage{float}
\usepackage{amsmath}
\usepackage{amssymb}
\usepackage{mathrsfs}
\usepackage{booktabs}
\usepackage{bbm}

\usepackage[pagebackref,breaklinks,colorlinks]{hyperref}

\ifCLASSINFOpdf
\else
\fi
\UseRawInputEncoding

\begin{document}
%
\title{2D Human Pose Estimation with Explicit Anatomical Keypoints Structure Constraints}
%
%
%

\author{Zhangjian ~Ji,~\IEEEmembership{Member,~IEEE,}
        Zilong ~Wang, Ming ~Zhang, Yapeng ~Chen and
        Yuhua ~Qian,~\IEEEmembership{Member,~IEEE,}
\thanks{Zhangjian Ji, Zilong Zhang, Ming Zhang and Yapeng Chen are with the Key Laboratory of Computational Intelligence
and Chinese Information Processing of Ministry of Education, School of Computer and Information Technology, Shanxi University, Taiyuan 030006, China E-mail: \{jizhangjian@sxu.edu.cn\}}
\thanks{Yuhua Qian is with the Institute of Big Data Science
and Industry, the Key Laboratory of Computational Intelligence and
Chinese Information Processing of Ministry of Education, Shanxi University,
Taiyuan 030006, Shanxi, China E-mail: \{ jinchengqyh@126.com \}}
}

%
%

\markboth{Journal of \LaTeX\ Class Files,~Vol.~14, No.~8, November~2022}%
{Shell \MakeLowercase{\textit{et al.}}: Bare Demo of IEEEtran.cls for IEEE Journals}
%



\maketitle

\begin{abstract}
Recently, human pose estimation mainly focuses on how to design a more effective and better deep network structure as human features extractor, and most designed feature extraction networks only introduce the position of each anatomical keypoint to guide their training process. However, we found that some human anatomical keypoints kept their topology invariance, which can help to localize them more accurately when detecting the keypoints on the feature map. But to the best of our knowledge, there is no literature that has specifically studied it. Thus, in this paper, we present a novel 2D human pose estimation method with explicit anatomical keypoints structure constraints, which introduces the topology constraint term that consisting of the differences between the distance and direction of the keypoint-to-keypoint and their groundtruth in the loss object. More importantly, our proposed model can be plugged in the most existing bottom-up or top-down human pose estimation methods and improve their performance. The extensive experiments on the benchmark dataset: COCO keypoint dataset, show that our methods perform favorably against the most existing bottom-up and top-down human pose estimation methods, especially for Lite-HRNet, when our model is plugged into it, its AP scores separately raise by 2.9\% and 3.3\% on COCO val2017 and test-dev2017 datasets.
\end{abstract}


%
\IEEEpeerreviewmaketitle

\section{Introduction}
%
%
%
%
\IEEEPARstart{H}{uman} pose estimation has been being a fundamental yet challenging research topic in the computer vision field, which aims at localizing the anatomical keypoints (e.g., eye, elbow, wrist, foot, etc.) or parts of each person from the image or video, and describes the human skeleton information by those ones. It has pervasive applications, including human action recognition, human-computer interaction, smart photo editing, pedestrian tracking, etc. Although human pose estimation has made great progress in the past decade, besides the difference of scale, appearance and number of people in each image, there is still many additional challenges (e.g., illumination variation, object occlusion, self-occlusion, etc.) to design a robust human pose estimation algorithm. Here, we mainly investigate how to improve the performance of existing human pose estimation methods when confronting the complex factors mentioned above.

At present, 2D human pose estimation algorithms can be categorized into the top-down and bottom-up methods. The top-down methods\cite{author27, author7, author28, author25, author31} first use the object detector (e.g., Faster-RCNN\cite{author1}, YOLO\cite{author2}, etc) to obtain the bounding box of each person instance from the image, and then crop the bounding box region of each person from the original image and resize them to approximately same scale and feed them to the keypoint detection network to predict all the keypoints of each person. Thus, they can simplify the multi-person pose estimation to single-person one and are generally less sensitive to the scale variance of persons. However, as such methods rely on a separate person detector and need to estimate the pose of each person individually, they are computationally expensive and not truly end-to-end system. By contrast, the bottom-up methods\cite{author21,author32,author24,author33,author18} primarily predict identity-free keypoints of all the persons in the input image by anatomical keypoints detection network, then group them into person instances. Both these methods generate a heatmap for each anatomical keypoint and locate its coordinates by using an argmax operation\cite{author3} on the heatmap. Generally speaking, the heatmap-based pose estimation methods are more accurate but not truly end-to-end system from the input image to human anatomical keypoints. Therefore, some researchers\cite{author5,author6} use pixel-wise keypoint regression approaches (e.g., CenterNet\cite{author4}) to directly predict the positions of human anatomical keypoints from the input image, which is more flexible and effective for various human pose estimation tasks, especially real-time anatomical keypoints estimation tasks on edge devices. However, the regression-based keypoint estimation methods are spatially inaccurate and its overall performance is worse than other two methods mentioned above. The main reason is that some challenging factors (e.g., occlusion, motion blur, truncation, etc) make the real labels of anatomical keypoints fuzzy and ambiguous, and the keypoint estimation approaches based on regression are more susceptible to these noise labels.

\begin{figure}
  \centering
  \includegraphics[width=7cm]{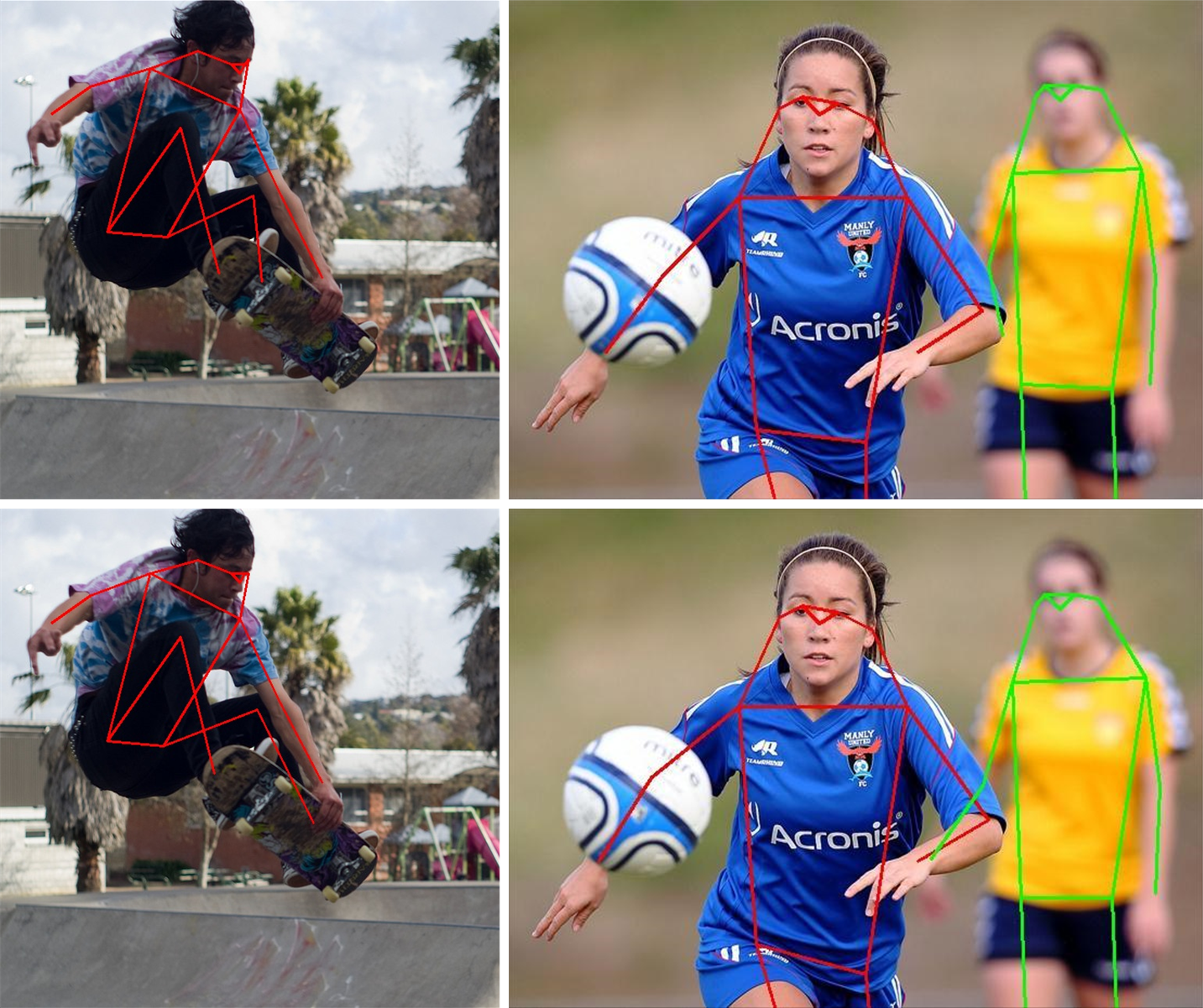}\\
  \caption{Illustration of human pose estimation. The first row gives the results of human pose estimation by Lite-HRNet\cite{author29}, and the second row shows the visualization of human pose estimation by Lite-HRNet with our proposed explicit anatomical keypoints structure constraints. It can be seen that the improved Lite-HRNet can locate the occluded human anatomical keypoints (e.g., the foot in the left image and the wrist in the right image) more accurately. }\label{fig_2}
\end{figure}

No matter which method is mentioned above, in previous literatures, most researchers focus on designing a more effective CNN structure to improve the feature extraction ability of the deep network model\cite{author7, author8,author9,author11,author13} or adopting some post-processing tricks to reduce the locating deviations of the predictive anatomical keypoints\cite{author14,author16,author17,author19,author20}, and they don't take into account that the mutual constraints between human anatomical keypoint pairs may affect the final positioning accuracy of each anatomical keypoint. In fact, some anatomical keypoints maintain their topology invariance (e.g., arm's length, the distance between two shoulders, etc) during human movement, which can help the keypoint detection network to locate human anatomical keypoints more accurately. If human anatomical keypoints are connected together, it is easy to find that the human pose is a graphical structure (see the Fig.\ref{fig_1}), the nodes and edges of which are respectively composed of human anatomical keypoints and the lines among them. As shown in Fig.\ref{fig_2}, the inherent mutual constraints between human keypoint pairs are beneficial to improve the predictive accuracy of human anatomical keypoints.

Therefore, in order to better explore how the mutual constraints among human joints affect the performance of human pose estimation, in this paper, we propose a novel 2D human pose estimation method with explicit anatomical keypoints structure constraints. The main contributions of this work are summarized as follows:
\begin{figure}
  \centering
  \includegraphics[width=3cm]{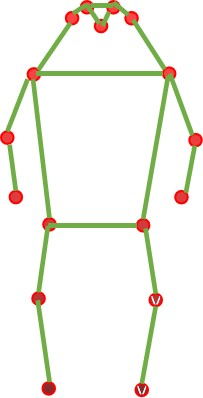}\\
  \caption{Illustration of human pose topology.}\label{fig_1}
\end{figure}
\begin{itemize}
  \item We first reveal the vital function of the structure constraints among human anatomical keypoints for human pose estimation model, and propose a plug-and-play explicit anatomical keypoints structure constraint model.
  \item We plug the proposed model into the existing popular human pose estimation methods, and the performance of these methods has been enhanced, especially for the Lite-HRNet, its AP scores separately raise by 2.9\% and 3.3\% on COCO val2017 and test-dev2017 datasets.
  \item Quantitative experiments and ablation study on the authoritative COCO dataset show that the proposed model and its each components are reasonable and effective.
\end{itemize}

 The remaining of this work is organized as follows: Section \ref{sec:rw} introduces the related works. In section \ref{sec:PM}, we elaborate the proposed human anatomical keypoints constraint model and how to plug it into the existing bottom-up and top-down methods. Section \ref{sec:exp} shows the ablation study and evaluation results on COCO keypoints datasets. Finally, conclusions are explained in Section \ref{sec:con}.
\section{Related works}\label{sec:rw}
\subsection{Bottom-up methods}
The bottom-up methods first predict identity-free keypoints of all the persons from the input image by human anatomical keypoints detection network, and then group them into person intances and connect the keypoints of each person instance to form the corresponding human pose based on the inherent linkage of human anatomical keypoints (e.g., left shoulder$\rightarrow$ left elbow in the Fig. \ref{fig_1}). OpenPose\cite{author21} firstly adopts a two-branch multi-stage network to predict the confidence maps of human anatomical keypoints location and the 2D vector field of human keypoint affinities which can encode the degree of association between human keypoint pairs, and then parse the anatomical keypoints belonging to the same person by the bipartite graph matching based on the part affinity fields. Newell \emph{et al.}\cite{author22} integrate the associative embedding with a stacked hourglass network\cite{author23}, which can simultaneously produce a detection heatmap and a tagging heatmap for each body joint and group body joints with similar tags into individual people, and achieved the optimal performance of the multi-person pose estimation until that year. Cheng \emph{et al.}\cite{author24} propose a HigherHRNet, which is the extension in HRNet (High resolution network)\cite{author25} and equipped with multi-resolution supervision for training and multi-resolution aggregation for inference. So it can solve the scale variation challenge in bottom-up multi-person pose estimation and locate keypoints more precisely, especially for small person. Although these bottom-up methods have good performance for human pose estimation, they are quite large and computationally expensive and not suitable for Internet-of-Things (IoT) applications requiring lightweight real-time multi-person pose estimation at the edge, next to the camera. Therefore, some researchers develop a number of lightweight bottom-up methods (e.g., EfficientHRNet\cite{author26}, etc) that are capable of real-time execution under constrained computational resources.

\subsection{Top-down methods}
The top-down methods firstly use an object detector to locate the bounding boxes of all the persons from the input image, and then crop the bounding box region of each person from the original image and perform single-person pose estimation on these cropping patches. So the core of the top-down methods focuses on how to better carry out single-person pose estimation on the cropping patches. For example, Papandreou \emph{et al.}\cite{author27} propose a simple, yet powerful, top-down human pose estimation approach consisting of two stages, which predicts the bounding box of each person from the input image by the FasterRCNN\cite{author1} and crops the corresponding bounding box regions in the first stage and  adopts a full convolutional ResNet\cite{author34} to predict activation heatmaps and offsets for each human anatomical keypoint in second stage. Chen \emph{et al.}\cite{author7} propose a cascaded pyramid network, which integrates global pyramid network (GlobalNet) and pyramid refined
network based on online hard keypoints mining (RefineNet) that are used to predict the simple and hard human anatomical keypoints, respectively. Mask R-CNN\cite{author10} directly adds a mask branch on the FasterRCNN\cite{author1} to predict $K$ mask, one for each of $K$ human keypoint types (e.g., left shoulder, right elbow). Bin \emph{et al.}\cite{author28} provide a simplest baseline method for the top-down human pose estimation, which inserts a few deconvolutional layers into the ResNet to expand the size of the low-resolution feature map so as to generate the high-resolution heatmaps for predicting human keypoints more precisely. Sun \emph{et al.}\cite{author25} propose a high-resolution network, HRNet for short, which has a distinctive parallel structure and always maintains high-resolution representations through the whole process. Due to its these two advantages, the keypoint heatmap predicted by HRNet is potentially more accurate and spatially more precise so that most human pose estimation methods choose it as the bone network. Similar to the bottom-up human pose estimation methods, to perform human pose estimation on the edge devices, some scholars also present a few lightweight top-down methods (e.g., Dite-HRNet\cite{author30}, Lite-HRNet\cite{author29}, etc), which apply the lightweight modules (e.g., Shuffle block in ShuffleNet\cite{author35}) to HRNet, greatly reducing their computation complexity and maintaining good performance.

\subsection{Other methods}
In addition to two human pose estimation methods mentioned above, there are other methods to directly regress the location of human anatomical keypoints from the input image. Although the overall performance of such methods is worse, they can truly achieve the end-to-end training and a few can also reach the accuracy of the top-down human pose estimation methods. For example, the CenterNet\cite{author4} directly regresses the location of human anatomical keypoints from the human center and the disentangled
keypoint regression (DEKR)\cite{author6} proposes a separate multi-branch regression scheme, each branch of which independently learns a representation with dedicated adaptive convolutions for each keypoint and regresses the position of the corresponding keypoint. Similar to anchor-based object detection, the PointSetNet\cite{author5} adopts a series of point-set anchors which follow the pose distribution of the training data to initialize the human pose in the input image and predicts the offsets of point-set anchors for real pose keypoints in the training process. As a good task-specific initialization, the point set can yield features that better facilitate keypoint localization.
\section{Proposed human anatomical keypoints constraints model}\label{sec:PM}
In this section, we specifically describes the proposed anatomical keypoints constraints model and how it can be plugged into the existing top-down and bottom-up human pose estimation methods.
\subsection{Generate the coordination set of human anatomical keypoints}\label{sec:l1cft}
Given an image $\textbf{I}\in\mathbb{R}^{w\times h}$, where $w$ and $h$ denote the width and height of image, human pose estimation aims to predict the coordination set $\emph{J}=\{(x,y)_{k,i}|k=1,2,\cdots,K, i=1,2,\cdots, N\}$ of all the human anatomical keypoints, where $K$ denotes the number of human anatomical keypoints (e.g., $K=17$ on the MS COCO keypoint dataset), and $N$ is the number of person instances in the image $\textbf{I}\in\mathbb{R}^{w\times h}$.

\begin{figure*}
  \centering
  \includegraphics[width=16cm]{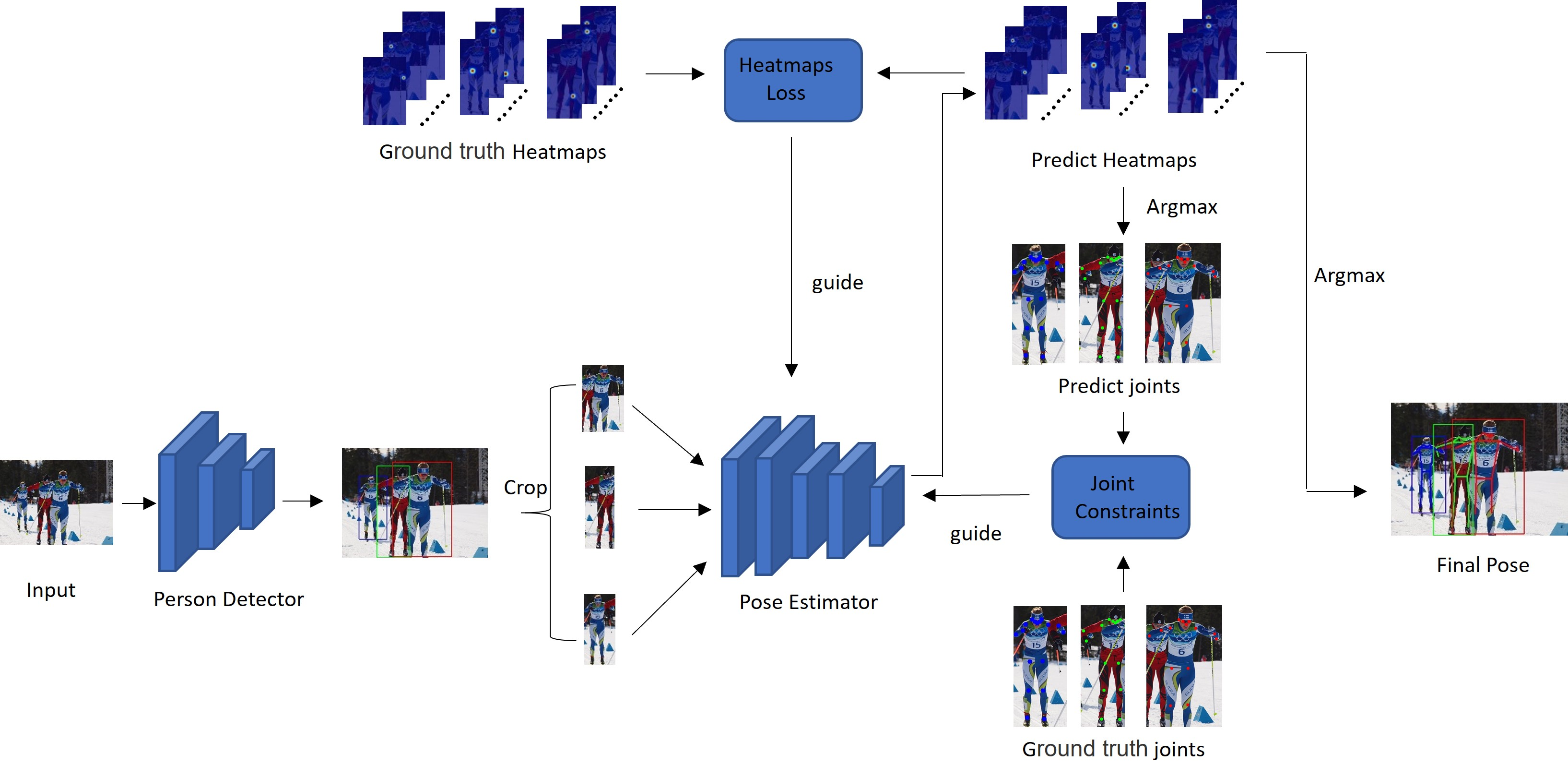}\\
  \caption{The overall work-flow of top-down human pose estimation methods with explicit keypoints structure constraints.}\label{fig_3}
\end{figure*}

The top-down human pose estimation methods first need to detect the bounding box of each person from the input image $\textbf{I}\in\mathbb{R}^{w\times h}$, and then crop the human patches $\emph{\textbf{P}}=\{\textbf{P}_{i}|i=1,2,\cdots,N\}$ from the original image $\textbf{I}\in\mathbb{R}^{w\times h}$ according to the detected bounding boxes and feed them into human keypoints detection network to generate their human anatomical keypoints heatmaps $\emph{\textbf{H}}=\{\textbf{H}_{i}\in\mathbb{R}^{w\times h\times K}|i=1,2,\cdots,N\}$. Finally, we obtain the coordination set $\emph{J}_{i}=\{(x,y)_{k}|k=1,2,\cdots,K\}, i=1,2,\cdots,N$ of human anatomical keypoints on each heatmap $\textbf{H}_{i},i=1,2,\cdots,N$ by an argmax operation\cite{author3} (see the Figure \ref{fig_3} in details).

\begin{figure*}
  \centering
  \includegraphics[width=16cm,height=8.5cm]{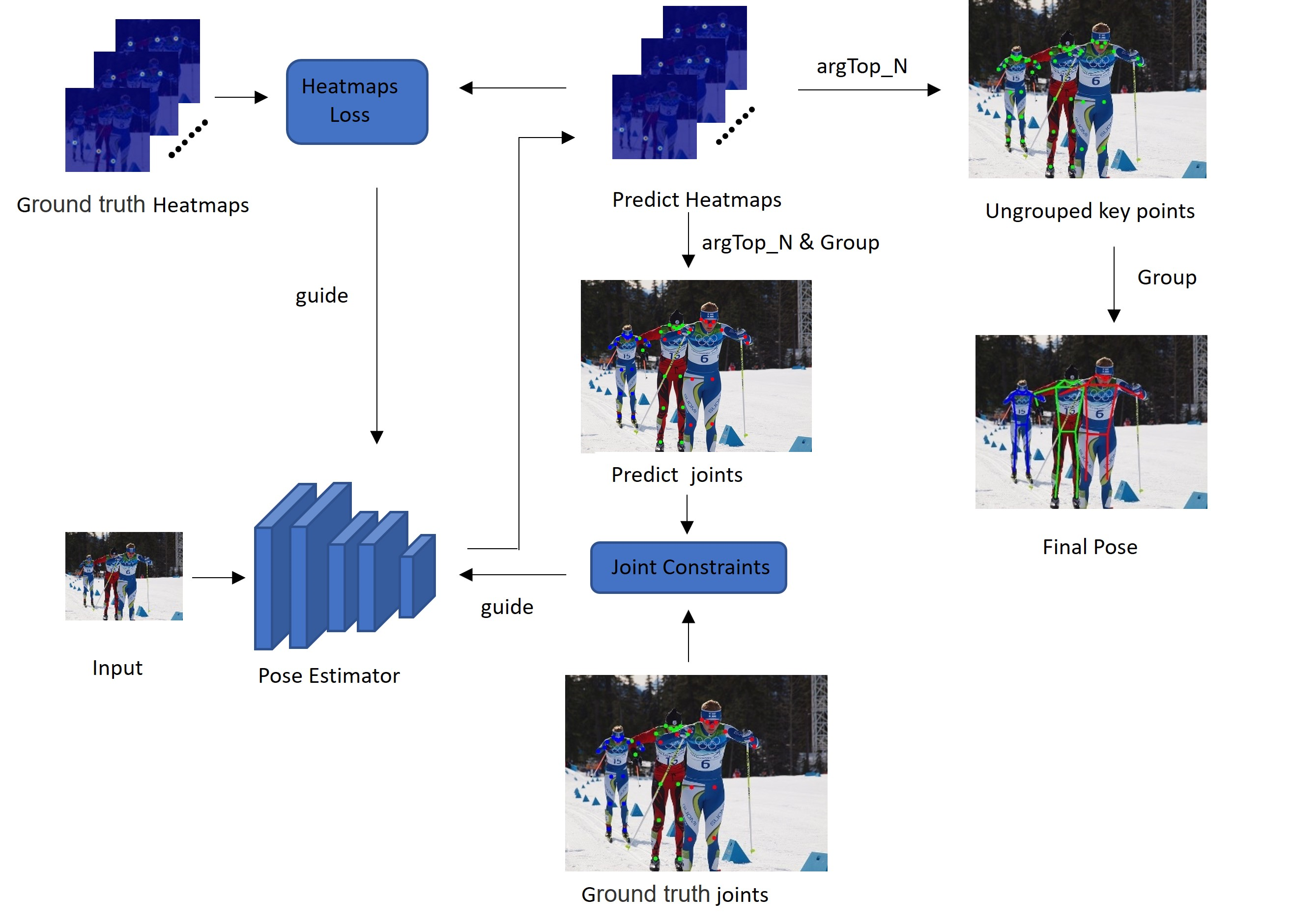}\\
  \caption{The overall work-flow of bottom-up human pose estimation methods with explicit keypoints structure constraints.}\label{fig_4}
\end{figure*}

Different from the top-down human pose estimation methods, the bottom-up ones directly adopt the human keypoints detection network to generate the heatmaps $\textbf{H}\in \mathbb{R}^{w\times h \times K}$ of all human keypoints from the original input image $\textbf{I}\in\mathbb{R}^{w\times h}$, where each heatmap $\textbf{H}_{k}\in \mathbb{R}^{w\times h},k=1,2,\cdots,K$ contains the identical class anatomical keypoint of all persons in the image. Next, the anatomical keypoints  of each heatmap $\textbf{H}_{k}\in \mathbb{R}^{w\times h},k=1,2,\cdots,K$ can be obtained by the following formula
\begin{equation}\label{eq1}
  \emph{J}_{k}=\arg \mathop{Top\_N}\limits_{(x,y)\in\Omega}(\textbf{H}_{k}(x,y)),
\end{equation}
where $\Omega=\{0,1,2,\cdots,w-1\}\times\{0,1,2,\cdots,h-1\}$, $Top\_N()$ describes the choosing the top $N$ maximal values on the heatmap, and $\emph{J}_{k}=\{(x,y)_{i}|i=1,2,\cdots,N\}$ denotes the coordination set of anatomical keypoints obtained from the heatmap $\textbf{H}_{k}$. Then, when all the heatmaps are done, the coordination set $J=\{(x,y)_{j}|j=1,2,\cdots,K\times N\}$of all the anatomical keypoints in the input image $\textbf{I}\in\mathbb{R}^{w\times h}$ can be obtained. However, the coordination set $J$ is not grouped, and not decide which anatomical keypoints belong to the identical person instance. Thus, we use associative embedding\cite{author22} to group the coordination set $J$ of all the anatomical keypoints by the following formula
\begin{equation}\label{eq2}
 J_{i}=group(J) \quad i=1,2,\cdots,N,
\end{equation}
where $J_{i}=\{(x,y)_{k}|k=1,2,\cdots,K\}$, and $group()$ denotes a grouping operation for all the anatomical keypoints. More intuitively, please see the Figure \ref{fig_4}.

However, for the regression-based human anatomical keypoints estimation methods, they can directly predict the coordinations set $\emph{J}_{i}=\{(x,y)_{k}|k=1,2,\cdots,K\}, i=1,2,\cdots,N$ of human anatomical keypoints from the input image.
\subsection{Human anatomical keypoints constraint model}
After obtaining the coordination set $\emph{J}=\{(x,y)_{k,i}|k=1,2,\cdots,K, i=1,2,\cdots, N\}$ of human anatomical keypoints, as shown in the Figure \ref{fig_1}, we only consider the structure constraints between human anatomical keypoints that have a connecting edge because these connecting edges keep invariance during the human movement (e.g., arms, legs, etc). And the distances among other anatomical keypoints without the inherent connecting edges are unpredictable due to the human motion. Thus, we first define an undirected graph $G=(V,E)$ that represents the topology of human anatomical keypoints, where $V$ is the equal to human anatomical keypoints set $J$, $E$ denotes the connecting edges set ($|E|=19$ in this paper) and $\textbf{e}_{k_{1},k_{2}}=(x_{k_{1}}-x_{k_{2}},y_{k_{1}}-y_{k_{2}})\in E, k_{1},k_{2}\in\{1,2,\cdots,K\}$, represents the edge between human anatomical keypoints $(x,y)_{k_{1}}$ and $(x,y)_{k_{2}}$ that have the connecting relation. Then, in order to better guide the training of keypoints detection network and improve the predictive accuracy of human anatomical keypoints, we can define the following human anatomical keypoints constraint model
\begin{equation}\label{eq3}
 \mathcal{L}_{cst}=\sum\limits_{k_{1},k_{2}}^{|E|^{}}(\lambda_{k_{1},k_{2}}\|\hat{\textbf{e}}_{k_{1},k_{2}}-\textbf{e}_{k_{1},k_{2}}\|^{2}+(1-\cos\theta_{k_{1},k_{2}})),
\end{equation}
where $\hat{\textbf{e}}_{k_{1},k_{2}}$ denotes the edge between the predicted human anatomical keypoints $(\hat{x},\hat{y})_{k_{1}}$ and $(\hat{x},\hat{y})_{k_{2}}$, $\textbf{e}_{k_{1},k_{2}}$ denotes the edge between the real human anatomical keypoints $(x,y)_{k_{1}}$ and $(x,y)_{k_{2}}$, $\cos\theta_{k_{1},k_{2}}$ is the cosine of the angle between the $\hat{\textbf{e}}_{k_{1},k_{2}}$ and $\textbf{e}_{k_{1},k_{2}}$ and $\lambda_{k_{1},k_{2}}$ is the weight that measure the contribution of the edge $\textbf{e}_{k_{1},k_{2}}$.

\begin{figure}
  \centering
  \includegraphics[width=4cm]{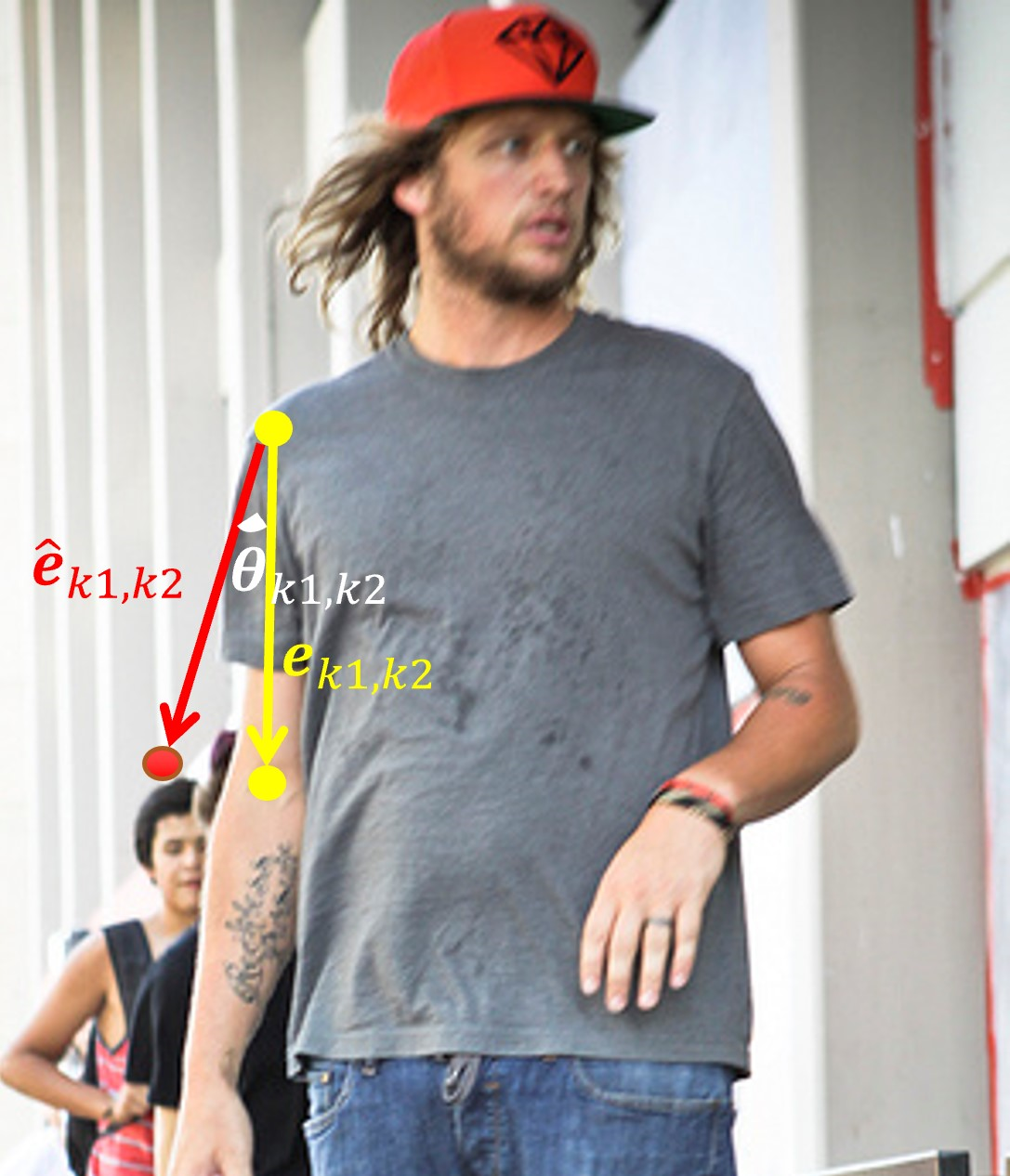}\\
  \caption{Illustration of human anatomical keypoints constraint.}\label{fig_5}
\end{figure}

\begin{figure}
  \centering
  \includegraphics[width=8cm]{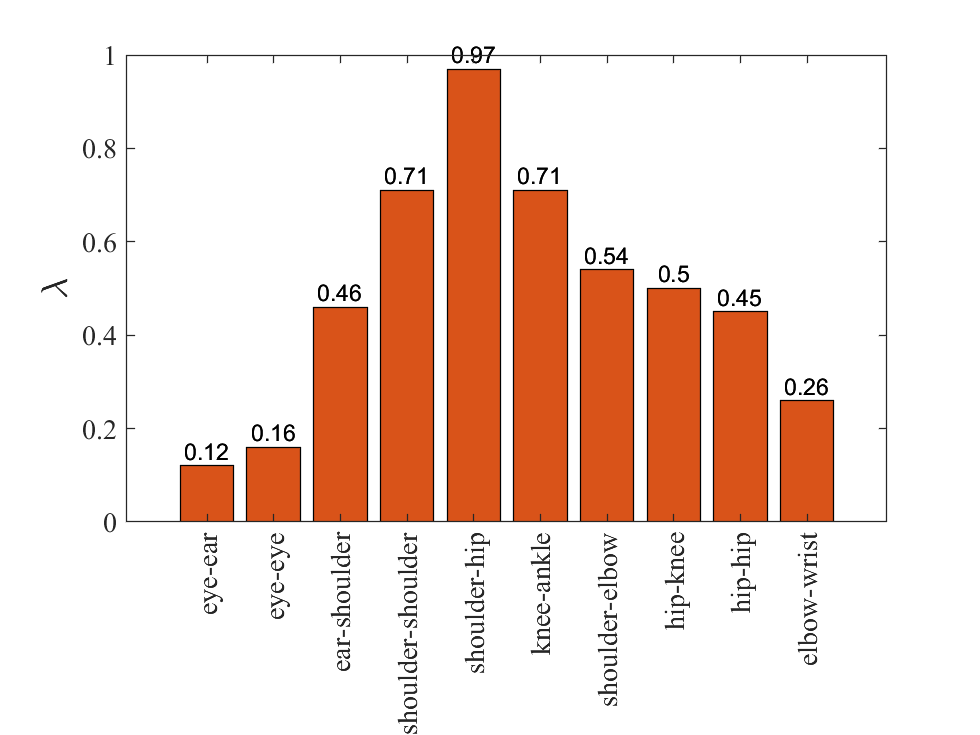}\\
  \caption{The $\lambda$ of the connecting edges shown in the Figure \ref{fig_1}. For the symmetric connecting edges, only show the weight $\lambda$ of one of them.}\label{fig_6}
\end{figure}

As shown in the Figure \ref{fig_5}, to minimize the discrepancy between the $\hat{\textbf{e}}_{k_{1},k_{2}}$ and $\textbf{e}_{k_{1},k_{2}}$, we need to minimize their length difference and angle simultaneously. The constraint model given in the Eq.(\ref{eq3}) can satisfy it and realize the structure constraint among human abatomical keypoints. In addition, to balance the contribution of the different $\textbf{e}_{k_{1},k_{2}}$ for the constraint model, we assign the different weight for them by the following formula
\begin{equation}\label{eq4}
\lambda_{k_{1},k_{2}}=\frac{|\bar{\textbf{e}}_{k_{1},k_{2}}|}{\max(\bar{\textbf{e}}_{k_{1},k_{2}})},
\end{equation}
where $\bar{\textbf{e}}_{k_{1},k_{2}}$ denotes the connecting edge of the average pose in the Figure \ref{fig_1}. Besides, in order to make the symmetric connecting edges (e.g., the edges from the left shoulder to the left elbow and from the right shoulder to the right elbow) own the same weights, we adopt their average weights as their weights and shown them in the Figure \ref{fig_6}.
\subsection{Loss function}
 In order to fairly compare the various previous human pose estimation methods and the improved ones that plugged our human anatomical keypoints constraint model into them, we strictly followed their basic architecture and only add our human keypoints constraint module to the orginal loss function. Thus, we can define a general loss function as follows
 \begin{equation}\label{eq5}
   \mathcal{L}=\mathcal{L}_{original}+\mathcal{L}_{cst},
 \end{equation}
where $\mathcal{L}_{cst}$ is the human anatomical keypoints constraint loss function by the Eq.(\ref{eq3}) and the definition of $\mathcal{L}_{original}$ depends on the selecting original human pose estimation methods. For the top-down and bottom-up human pose estimation methods, $\mathcal{L}_{original}$ is usually a heatmap loss function based on the pixel-wise MSE minimization\cite{author24,author25}. And for the regression-based human pose estimation methods, their $\mathcal{L}_{original}$ is composed of the classification and regression losses. Take PointSetNet\cite{author5} for example, its loss function is defined as
 \begin{equation}\label{eq6}
 \begin{split}
 \mathcal{L}_{original}=&\frac{1}{N_{pos}}\sum\limits_{x,y}\mathcal{L}_{cls}(p_{x,y},c_{x,y}^{*})+\frac{\beta}{N_{pos}}\\
 &\sum\limits_{x,y}\mathbbm{1}_{\{c_{x,y}^{*}\}>0}\mathcal{L}_{reg}(t_{x,y},t_{x,y}^{*}),
 \end{split}
 \end{equation}
where $\mathcal{L}_{cls}$ is the focal loss in \cite{author34} and $\mathcal{L}_{reg}$ is the $\ell_{1}$ loss for shape regression. $c_{x,y}^{*}$ and $t_{x,y}^{*}$ respectively represent classification and regression targets. $N_{pos}$ denotes the number of positive sample and $\beta$ is the balance weight, which is set to $10.0$. $\mathbbm{1}_{\{c_{x,y}^{*}\}>0}$ is an indicator function, being 1 if $\{c_{x,y}^{*}\}>0$ and 0 otherwise.
\section{Experiments}\label{sec:exp}
In this section, we first introduce the benchmark datasets and evaluation metrics used in our experiments. Next, we describe the implementation details. Then, we conduct ablation study to analyze the effect of parameter $\lambda$ and each constraint term in our proposed human keypoints constraint model. Finally, we compare some lightweight state-of-the-art human pose estimation methods with their extensions which are plugged our huamn keypoints constraint model in them.
\subsection{Experimental setup}
\noindent\textbf{Datasets.} Plugging our proposed human keypoints constraint model into the various existing human pose estimation methods, we comprehensively evaluate its effectiveness for improving the performance of their human pose estimation on the MS COCO keypoint detection dataset (in 2017)\cite{author36} which contains over 200K images and 150K person instances labels with 17 keypoints and is divided into \emph{train/val/test-dev} sets respectively. All the experiments in this paper are trained only on the \emph{train2017} set, including 57K images and 150K person instances annotated with 19 keypoints. We report the results on the \emph{val2017} set for ablation studies and also give the comparison results on the \emph{val2017} and \emph{test-dev2017} sets.

\noindent\textbf{Evaluation metric.} We follow the standard evaluation metric (Object Keypoint Similarity--OKS\cite{author24, author25})\footnote{\url{http://cocodataset.org/\#keypoints-eval}} to measure the performance of each human pose estimation method on the COCO keypoint dataset. Thus, for each human pose estimation methods, in this paper, we report their standard average precision and average recall scores with different thresholds and different object sizes: $AP$ (the mean of $AP$ scores at $OKS = \{0.50, 0.55,\cdots. 0.90,0.95\}$), $AP^{50}$ (AP at $OKS=0.50$), $AP^{75}$, $AP^{M}$ for medium objects, $AP^{L}$ for large objects and $AR^{50}$ (the recall scores at $OKS=0.50$).
\subsection{Implementation details}
In this paper, all the experiments are run on a single NVIDIA A100 GPU with 40G RAM. In order to fairly compare the performance of each human pose estimation method with or without our proposed keypoints constraint model, we follow the training strategy of each human pose estimation method in the original papers and adopt the same number of epochs and the same learning rate to retrain them and their extensions (add our proposed keypoints constraint model on them) on COCO \emph{train2017} dataset by our equipment. For example, the PointSetAnchor\cite{author5} is trained by Adam optimizer\cite{author37}. The base learning rate is set to $1e-4$ and dropped to $1e-5$ and $1e-6$ at the $80th$ and $90th$ epoch respectively. There are 100 epochs in total. Samples with OKS higher than 0.5 and lower than 0.4 are defined as positive samples and negative samples, respectively.
\subsection{Ablation study}
On COCO $val2017$ dataset, we perform a number of ablation experiments to validate the effectiveness of our proposed keypoints constraint model from different aspects, which use Lite-HRNet as an example.

\noindent\textbf{Effect of parameter $\lambda$.} Intuitively, for our proposed keypoints constraint model, some connecting edges between keypoints (e.g., hand$\rightarrow$elbow, foot$\rightarrow$knee, etc) should have the significant effects on the accuracy of human pose estimation, and other are insignificant (e.g., nose$\rightarrow$eye, eye$\rightarrow$eye, etc). To verify this conclusion, we design three different calculation schemes for the $\lambda$ of the formula (\ref{eq3}), i.e., the scheme-1 is that each $\lambda_{k_{1},k_{2}}, k_{1},k_{2}\in\{1,2,\cdots,K\}$ is set to 1, the scheme-2 is $\lambda_{k_{1},k_{2}}=1-\frac{|\bar{\textbf{e}}_{k_{1},k_{2}}|}{\max(\bar{\textbf{e}}_{k_{1},k_{2}})}$ and the scheme-2 is that each $\lambda_{k_{1},k_{2}}$ is calculated by the formula (\ref{eq4}). For the scheme-3, the shortest edge is assigned the smallest weight (e.g., as shown in the Figure \ref{fig_6}, the weight $\lambda$ of eye$\rightarrow$eye is 0.12.) and vice versa. Contrary to the scheme-3, the shortest edge is assigned the largest weight in the scheme-2. Table \ref{tab:ablation-1} reports their comparison results on the COCO $val2017$ dataset, where the first row is the results of the base Lite-HRNet\cite{author29} and other three rows give the results of the different $\lambda$ calculation schemes after adding our proposed keypoints constraint model in Lite-HRNet. Seeing from the Table \ref{tab:ablation-1}, for three different $\lambda$ calculation schemes, we find that not only do their $AP$ scores raise by 0.8\%,1.6\% and 2.9\% than the base Lite-HRNet respectively but other metrics are also higher than its,  which suggests that our proposed keypoints constraint model can help to improve the performance of the existing human pose estimation methods (e.g., Lite-HRNet). The $AP$ score of the scheme-1 is lower by 0.8\% and 2.1\% than the scheme-2 and scheme-3 respectively, which proves that the importance of the different length's connecting edges is not the same in our proposed keypoints constraint model and assigning the different weights to them is beneficial to further improve the performance of ours on human pose estimation. Furthermore, seeing the scheme-2 and scheme-3 in Table \ref{tab:ablation-1}, we also notice that all the evaluation metrics of the scheme-3 are superior to those of the scheme-2, which demonstrates that the more longer connecting edges (e.g., hand$\rightarrow$elbow, foot$\rightarrow$knee, etc) play a more important role in boosting the performance of human pose estimation for our proposed keypoints constraint model.
\begin{table*}
  \caption{Ablation study of three different $\lambda$ calculation schemes for Lite-HRNet added our proposed keypoints constraint model on the COCO $val2017$ dataset. The best results are highlighted by the bold.}
  \centering
  \begin{tabular}{cccccccc}
    \toprule
     $\lambda$& backbone & Input size & AP(\%) &$AP^{50}$(\%)&$AP^{75}$(\%)&$AP^{M}$(\%)&$AP^{L}$(\%)\\
    \midrule
    -- &Lite-HRNet-18 &$256\times192$ &63.6 &86.2 &71.2 &60.8&69.4\\
    scheme-1 &Lite-HRNet-18 &$256\times192$ &64.4 &86.8 &72.2 &61.8&70.1\\
    scheme-2 &Lite-HRNet-18 &$256\times192$ &65.2 &87.1 &73.3 &62.4&70.9\\
    scheme-3 &Lite-HRNet-18 &$256\times192$ &\textbf{66.5} &\textbf{87.4} &\textbf{74.7} &\textbf{63.8}&\textbf{72.3}\\
    \bottomrule
  \end{tabular}
  \label{tab:ablation-1}
\end{table*}

\noindent\textbf{Effect of each term in our keypoint constraint model.} Seeing the formula (\ref{eq3}), it is known that our proposed human keypoints constraint model includes two terms: the length and angle constraints, i.e., the first and second terms of the formula (\ref{eq3}), which can characterize whether the vectors connecting the predicted any two human keypoints with the inherent connecting edge are equal to the ones composed of their real values. In order to fully analyze their contributions to improve the performance of human pose estimation methods, taking the Lite-HRNet as the baseline, we further conduct some ablation studies on COCO $val2017$ dataset. The corresponding experimental results are reported in Table \ref{tab:ablation-2}. Seeing from the Table \ref{tab:ablation-2}, we observe that only length constraint or angle constraint term in our proposed keypoints constraint model is added into the Lite-HRNet and all the evaluation metrics obtains the significant improvement (e.g., its AP score raises by 2.6\%), which illustrates that they are both beneficial to enhance the performance of human pose estimation methods. When the length constraint and angel constraint terms of our proposed keypoints constraint model are both added into the Lite-HRNet, it obtains a gain of 2.9\% in AP score compared to the baseline (Lite-HRNet), even 0.3\% higher than the one using one of the two constraint terms, which further demonstrates that two constraint terms of our proposed keypoints constraint model are both very important for boosting the performance of human pose estimation methods.
\begin{table*}
  \caption{Influence of two different constraint terms ( see the formula (\ref{eq3}).) for Lite-HRNet added our proposed keypoints constraint model on the COCO $val2017$ dataset. The best results are highlighted by the bold.}
  \centering
  \resizebox{1.995\columnwidth}{!}{
  \begin{tabular}{ccccccccc}
    \toprule
     Length constraint & Angle constraint & backbone & Input size & AP(\%) &$AP^{50}$(\%)&$AP^{75}$(\%)&$AP^{M}$(\%)&$AP^{L}$(\%)\\
    \midrule
     & &Lite-HRNet-18 &$256\times192$ &63.6 &86.2 &71.2 &60.8&69.4\\
    \checkmark& &Lite-HRNet-18 &$256\times192$ &66.2 &87.3 &73.7 &63.4&72.1\\
    & \checkmark&Lite-HRNet-18 &$256\times192$ &66.2 &87.3 &74.4 &63.6&72.0\\
    \checkmark&\checkmark &Lite-HRNet-18 &$256\times192$ &\textbf{66.5} &\textbf{87.4} &\textbf{74.7} &\textbf{63.8}&\textbf{72.3}\\
    \bottomrule
  \end{tabular}
  }
  \label{tab:ablation-2}
\end{table*}
\begin{figure}[!h]
  \centering
  \includegraphics[width=7.5cm,height=6cm]{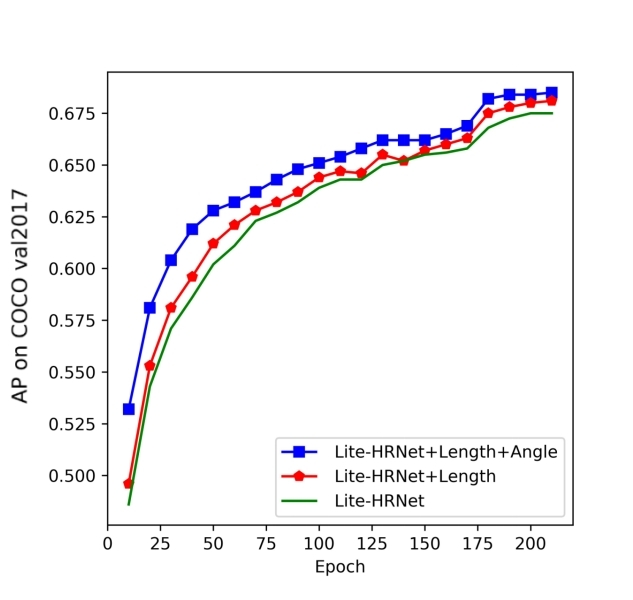}\\
  \caption{Curve of the AP scores of three different methods (Lite-HRNet, Lite-HRNet+Length and Lite-HRNet+Length+Angle) on COCO $val2017$ dataset in the training process. Here, Lite-HRNet+Length denotes that only length constraint term of our proposed keypoints constraint model is added into the Lite-HRNet and Lite-HRNet+Length+Angle represents that the Lite-HRNet contains our human keypoints constraint model defined in the formula (\ref{eq3}).}\label{fig_7}
\end{figure}
In addition, to analyze the influence of each component of our proposed keypoints constraint model on the convergence rate, in the Figure \ref{fig_7}, we plot the curves of AP scores of three methods (Lite-HRNet, Lite-HRNet+Length and Lite-HRNet+Length+Angle) on the different epochs. Seeing the curves of AP scores in the Figure \ref{fig_7}, we observe that the Lite-HRNet+Length+Angle ranks No.1 and Lite-HRNet+Length is second on the AP scores and convergence rate, which shows that our proposed human keypoints constraint model not only improve the performance of Lite-HRNet but also make it converge faster in the training process.
\subsection{Results}
Due to the limited computing power (only one NVDIA A100 GPU), in this subsection, to comprehensively validate the effectiveness of our proposed human keypoints constraint model, we select some lightweight human pose estimation methods (e.g., Lite-HRNet\cite{author29}, Dite-HRNet\cite{author30}, etc) and compare with those methods adding our proposed human keypoints constraint model on COCO \emph{val2017} and \emph{test-dev2017} datasets.

\noindent\textbf{COCO \emph{val2017}.} As shown in the Table \ref{tab:val}, we compare some state-of-the-art lightweight human pose estimation methods with their extensions which plugs our proposed human keypoints constraint model into the original methods. Seeing from the Table \ref{tab:val}, we find that any original human pose estimation methods obtain some performance gains on COCO \emph{val2017} dataset after they are added our human keypoints constraint model. For the single-stage methods PointSetAnchor\cite{author5} and DEKR\cite{author6}, their extensions are 0.7\% and 0.9\% higher than their original methods in terms of the AP scores, respectively, the extension of the EfficientHRNet\cite{author26} is superior to itself by 0.6\% in AP score, and for Lite-HRNet\cite{author29}, Dite-HRNet\cite{author30} and its variant, their extensions separately obtain a gain of 0.3\%, 0.1\% and 2.9\% in terms of AP scores. In the Table \ref{tab:val}, we also observed a meaningful result, i.e., for the same type human pose estimation methods, the worse the performance of the original methods is and the performance gain of their extensions is more significant, e.g., the performance of Lite-HRNet is the worst among three top-down human pose estimation methods in Table \ref{tab:val} but the performance gain of its extension (Lite-HRNet*) is the most significant among that of their extensions and raise by 2.8\% than that of the worst one (Dite-HRNet*-30) of them. For the different type human pose estimation methods, the AP scores of the DEKR and Dite-HRNet-30 have no significant difference, with only a 0.2\% gap. But the performance gain of the DEKR* is 0.8\% higher than that of Dite-HRNet*-30, which proves that our human keypoints constraint model is more effective for improving the performance of the single-stage human pose estimation methods, because they are spatially inaccuracy and are inferior to the bottom-up and top-down human pose estimation methods under the case of the same input size and backbone. In addition, we also notice that the Dite-HRNet-18 obtains a gain of 1.9\% than the Lite-HRNet-18 in terms of AP score, and the reason is that it can acquire the abundant contextual information\cite{author30}. However, the extension of the Lite-HRNet-18 outperforms the Dite-HRNet-18 by 1\% in terms of AP score, which shows that our human keypoints constraint model can bridge the performance gap between the Lite-HRNet-18 and Dite-HRNet-18, even making the Lite-HRNet-18 perform better.
\begin{table*}
  \caption{Comparison results on the COCO \emph{val2017} dataset. * denotes that the corresponding human pose estimation method is added our human keypoints constraint model. The best results of each method are highlighted by the bold.}
  \centering
  \resizebox{1.995\columnwidth}{!}{
  \begin{tabular}{ccccccccc}
    \toprule
     Methods & Backbone & Input size & AP(\%) &$AP^{50}$(\%)&$AP^{75}$(\%)&$AP^{M}$(\%)&$AP^{L}$(\%)&$AR^{50}$(\%)\\
     \midrule
     \multicolumn{9}{c}{Single-stage methods (regression-based methods)}\\
    \midrule
     PointSetAnchor\cite{author5}&HRNet-W32&$640\times640$&69.1 &87.7 &75.6 &\textbf{67.4} &74.1&--\\
     PointSetAnchor*&HRNet-W32&$640\times640$&\textbf{69.8} &\textbf{88.2} &\textbf{76.8} &67.3 &\textbf{75.8}&--\\
     DEKR\cite{author6}&HRNet-W32&$640\times640$&67.6 &86.6 &74.5 &61.4 &77.6&72.7\\
     DEKR*&HRNet-W32&$640\times640$&\textbf{68.5} &\textbf{87.4} &\textbf{74.8} &\textbf{62.8} &\textbf{77.9}&\textbf{73.5}\\
     \midrule
     \multicolumn{9}{c}{Bottom-up methods}\\
     \midrule
    EfficientHRNet\cite{author26}&EfficientHRNet-H\_2 &$448\times448$ & 57.0&79.5 &61.6 &51.8 &64.8&64.9\\
    EfficientHRNet*&EfficientHRNet-H\_2 &$448\times448$ & \textbf{57.6}&\textbf{79.8} &\textbf{62.6} &\textbf{52.3} &\textbf{65.2}&\textbf{65.0}\\
     \midrule
     \multicolumn{9}{c}{Top-down methods}\\
     \midrule
     Dite-HRNet\cite{author30}&Dite-HRNet-18 &$256\times192$ & 65.5&87.0 &73.2 &62.9 &71.1&71.6\\
     Dite-HRNet*&Dite-HRNet-18 &$256\times192$ & \textbf{65.8}&\textbf{87.3} &\textbf{74.0} &\textbf{63.3} &\textbf{71.5}&\textbf{72.2}\\
     Dite-HRNet\cite{author30}&Dite-HRNet-30 &$256\times192$ & 67.8&88.1 &75.8 &64.8 &73.9&73.8\\
     Dite-HRNet*&Dite-HRNet-30 &$256\times192$ & \textbf{67.9}&\textbf{90.5} &\textbf{76.0} &\textbf{66.1} &73.9&\textbf{73.9}\\
     Lite-HRNet\cite{author29}&Lite-HRNet-18 &$256\times192$ & 63.6&86.2 &71.2 &60.8 &69.4&70.0\\
     Lite-HRNet*&Lite-HRNet-18 &$256\times192$ & \textbf{66.5}&\textbf{87.4} &\textbf{74.7} &\textbf{63.8} &\textbf{72.3}&\textbf{72.8}\\
    \bottomrule
  \end{tabular}
  }
  \label{tab:val}
\end{table*}

\begin{figure*}
  \centering
  \includegraphics[width=17.5cm,height=6cm]{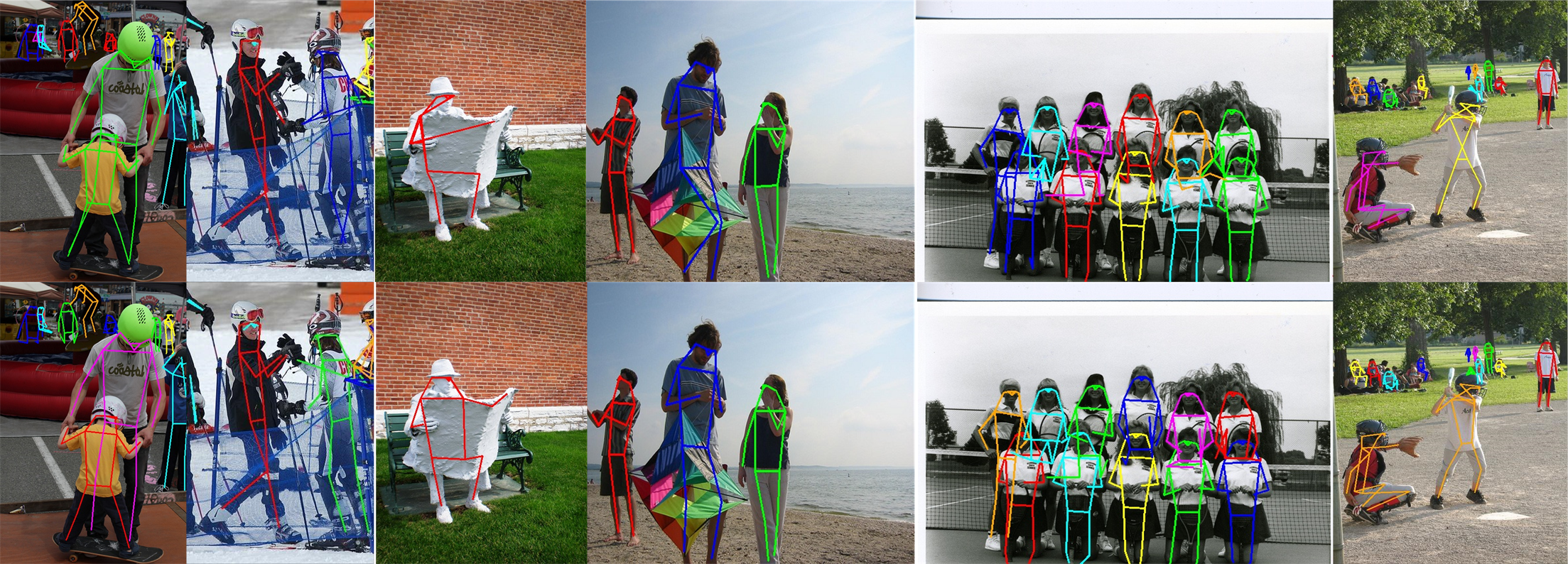}\\
  \caption{Qualitative results of some example images on COCO \emph{test-dev2017} dataset: containing view and appearance changing, occlusion, multiple persons, etc. The top row is the visual results of the Lite-HRNet\cite{author29}, and the bottom row is the visual results of the its extension (Lite-HRNet*) which plugs our proposed human anatomical keypoints constraint model.}\label{fig_8}
\end{figure*}

\begin{table*}
  \caption{Comparison results on the COCO \emph{test-dev2017} dataset. * denotes that the corresponding human pose estimation method is added our human keypoints constraint model. The best results of each method are highlighted by the bold.}
  \centering
  \resizebox{1.995\columnwidth}{!}{
  \begin{tabular}{ccccccccc}
    \toprule
     Methods & Backbone & Input size & AP(\%) &$AP^{50}$(\%)&$AP^{75}$(\%)&$AP^{M}$(\%)&$AP^{L}$(\%)&$AR^{50}$(\%)\\
     \midrule
     \multicolumn{9}{c}{Single-stage methods (regression-based methods)}\\
    \midrule
     DEKR\cite{author6}&HRNet-W32&$640\times640$&66.8 &87.8 &74.1 &60.8 &76.1&72.1\\
     DEKR*&HRNet-W32&$640\times640$&\textbf{67.7} &\textbf{88.3} &\textbf{74.6} &\textbf{62.0} &\textbf{76.5}&\textbf{72.8}\\
     \midrule
     \multicolumn{9}{c}{Bottom-up methods}\\
     \midrule
    EfficientHRNet\cite{author26}&EfficientHRNet-H\_2 &$448\times448$ & 56.6&80.8 &61.5 &51.7 &63.1&64.5\\
    EfficientHRNet*&EfficientHRNet-H\_2 &$448\times448$ & \textbf{57.1}&\textbf{81.2} &\textbf{62.3} &\textbf{52.3} &\textbf{63.4}&\textbf{64.7}\\
     \midrule
     \multicolumn{9}{c}{Top-down methods}\\
     \midrule
     Dite-HRNet\cite{author30}&Dite-HRNet-18 &$256\times192$ & 65.0&89.0 &73.0 &62.6 &69.8&71.1\\
     Dite-HRNet*&Dite-HRNet-18 &$256\times192$ & \textbf{65.3}&\textbf{89.2} &\textbf{73.6} &\textbf{63.0} &\textbf{70.2}&\textbf{71.4}\\
     Dite-HRNet\cite{author30}&Dite-HRNet-30 &$256\times192$ & 67.2&89.9 &75.4 &64.6 &72.1&73.0\\
     Dite-HRNet*&Dite-HRNet-30 &$256\times192$ & \textbf{67.3}&\textbf{90.3} &\textbf{75.6} &\textbf{65.8} &\textbf{72.2}&\textbf{73.1}\\
     Lite-HRNet\cite{author29}&Lite-HRNet-18 &$256\times192$ & 62.8&88.3 &70.3 &60.4 &67.7&69.1\\
     Lite-HRNet*&Lite-HRNet-18 &$256\times192$ & \textbf{66.1}&\textbf{89.6} &\textbf{74.2} &\textbf{63.6} &\textbf{70.9}&\textbf{72.2}\\
    \bottomrule
  \end{tabular}
  }
  \label{tab:test}
\end{table*}
\noindent\textbf{COCO \emph{test-dev2017}.} Table \ref{tab:test} reports the comparison results of some state-of-the-art lightweight human pose estimation methods and their extensions on COCO \emph{test-dev2017} dataset. We find that all the evaluation metrics of their extensions obtain the improvement, especially for the extension of the Lite-HRNet, its gain reaches 3.3\% in terms of AP score, which is consistent with the results of Table \ref{tab:val}. In addition, Figure\ref{fig_8} shows some visual results of the Lite-HRNet and its extension on COCO \emph{test-dev2017} dataset. We observe that the extension of Lite-HRNet can locate those occluded human anatomical keypoints more accurately than itself.

\section{Conclusion}\label{sec:con}
In this paper, we first reveal the vital function of human keypoints structure constraint for improving the performance of human pose estimation methods, and propose a plug-and-play explicit human keypoints constraint model. Then, we plug it into some state-of-the-art lightweight human pose estimation methods, which doesn't need to change their the basic architecture and hardly increase their training cost. The experimental results on COCO \emph{val2017} and \emph{test-dev2017} datasets demonstrate that our human keypoints constraint model can improve the performance of the existing human pose estimation methods, especially for the Lite-HRNet, its AP scores can raise by 2.9\% and 3.3\%, respectively. Although only given the results of some lightweight networks in this paper, in fact, our human keypoints constraint model can be also applied to other complex human pose estimation networks.


%

%

\section*{Acknowledgment}
This work is supported by the National Natural Science Foundation of China (No.61602288), Fundamental Research Program of Shanxi Province (No.20210302123443), National Key Research and Development Program of China
(No.2020AAA0106100), and the 1331 project of Shanxi Province. The authors also would like to thank the anonymous reviewers for their valuable suggestions.

\ifCLASSOPTIONcaptionsoff
  \newpage
\fi



%
{\small
\bibliographystyle{IEEEtran}
\bibliography{referenceBib}
}
%
%

%

\begin{IEEEbiography}[{\includegraphics[width=1in,height=1.25in,clip,keepaspectratio]{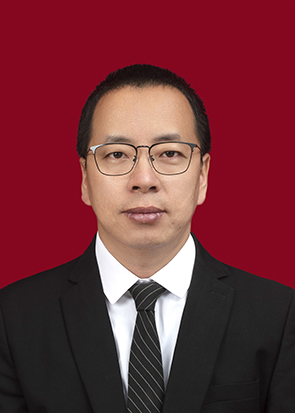}}]{Zhangjian Ji}
received the B.S. degree from Wuhan University, China, in 2007, the M.E. degrees from the Institute of Geodesy and Geophysics, Chinese
Academy of Sciences (CAS), China, in 2010, and the Ph.D. degree in the University of Chinese Academy of Sciences, China, in 2015.

He is currently an Associate Professor at the School of Computer and Information Technology, Shanxi University, Taiyuan, China. His research interests include computer vision, pattern recognition, machine learning and human-computer interaction.
\end{IEEEbiography}
\begin{IEEEbiography}[{\includegraphics[width=1in,height=1.25in,clip,keepaspectratio]{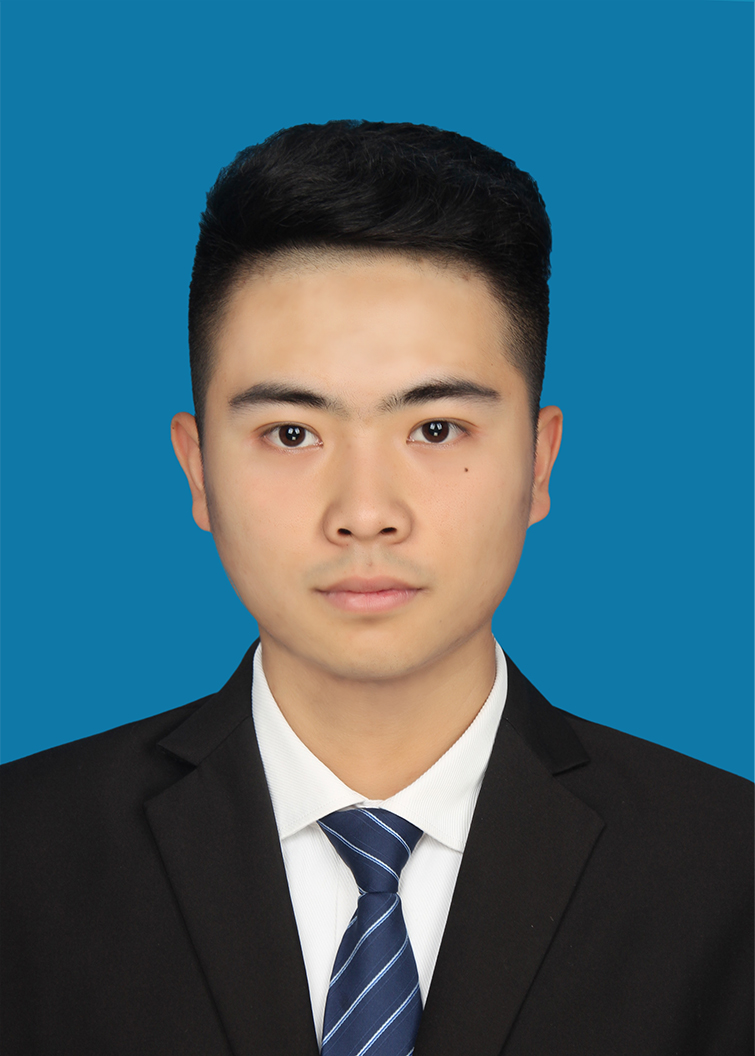}}]{Zilong Wang}
received the B.S. degree in engineering from Xiangtan University, China, in 2020. He is currently pursuing the M.E. degrees in Computer technologies from Shanxi university, China.

His research interests include computer vision, human-computer interaction, etc.
\end{IEEEbiography}
\begin{IEEEbiography}[{\includegraphics[width=1in,height=1.25in,clip,keepaspectratio]{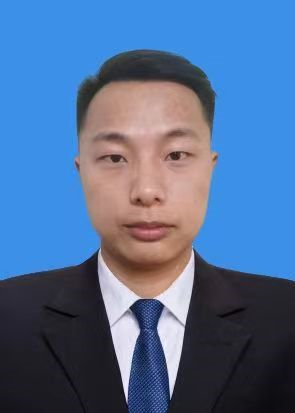}}]{ Ming Zhang}
received the B.S. degree in engineering from Henan Institute of Science and Technology, China, in 2020. He is currently pursuing the M.E. degrees in Computer technologies from Shanxi university, China.

His research interests include computer vision, pattern recognition, etc.
\end{IEEEbiography}
\begin{IEEEbiography}[{\includegraphics[width=1in,height=1.25in,clip,keepaspectratio]{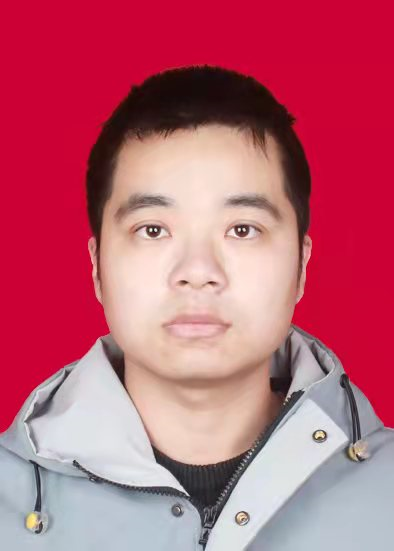}}]{Yapeng Chen}
received the B.S. degree in science from Shanxi Agricultural University, China, in 2020. He is currently pursuing the M.E. degrees in Computer technologies from Shanxi university, China.

His research interests include computer vision, human-computer interaction, etc.
\end{IEEEbiography}
\begin{IEEEbiography}[{\includegraphics[width=1in,height=1.25in,clip,keepaspectratio]{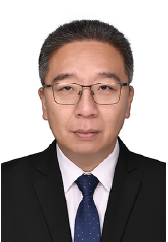}}]{Yuhua Qian}
is a professor and Ph.D. supervisor of Key Laboratory of Computational Intelligence and Chinese Information Processing of Ministry
of Education, China. He received the M.S. degree and the PhD degree in Computers with applications at Shanxi University in 2005 and 2011, respectively. He is best known for multigranulation rough sets in learning from categorical data and granular computing. He is actively pursuing research in pattern recognition, feature selection, rough set theory, granular computing and artificial intelligence. He has published more than 100 articles on these topics in international journals. He served on the Editorial Board of the International Journal of Knowledge-Based Organizations and Artificial Intelligence Research. He has served as the Program Chair or Special Issue Chair of the Conference on Rough Sets and Knowledge
Technology, the Joint Rough Set Symposium, and the International Conference on Intelligent Computing, etc., and also PC Members of many machine learning, data mining conferences.
\end{IEEEbiography}
%
%




\end{document}